\documentclass[draftclasofoot, onecolumn]{IEEEtran}
\IEEEoverridecommandlockouts
\usepackage[noadjust]{cite}
\usepackage[bookmarks=true,breaklinks=true,colorlinks,linkcolor=red,citecolor=blue,urlcolor=BlueViolet]{hyperref}
\usepackage{multirow}
\usepackage{amsmath,amssymb,amsfonts}
\usepackage[ruled,vlined]{algorithm2e}
\newcommand{\mycommentstyle}[1]{\color[HTML]{0671b9}{#1}}
\SetKwComment{Comment}{\mycommentstyle{// }}{}
\usepackage{graphicx}
\usepackage{threeparttable}
\usepackage{subfigure}
\usepackage{textcomp}
\usepackage{xcolor}
\def\BibTeX{{\rm B\kern-.05em{\sc i\kern-.025em b}\kern-.08em
    T\kern-.1667em\lower.7ex\hbox{E}\kern-.125emX}}
\usepackage{booktabs}
\usepackage{tablefootnote}
\usepackage{blindtext}
\usepackage{float}
\usepackage{algorithmic}
\usepackage{textcomp}
\usepackage{booktabs}

\begin{document}

\title{An Adaptive Latent Factorization of Tensors Model for Embedding Dynamic Communication Network}

\author{
    Xin Liao\textsuperscript{*} \quad Qicong Hu\textsuperscript{*} \quad Peng Tang\textsuperscript{*}\thanks{\textsuperscript{*}College of Computer and Information Science, Southwest University, Chongqing, China (lxchat26@gmail.com, mcshqc98@gmail.com, tangpeng.cn@outlook.com)}
}

\maketitle

\begin{abstract}
The \textit{Dynamic Communication Network} (DCN) describes the interactions over time among various communication nodes, and it is widely used in Big-data applications as a data source. As the number of communication nodes increases and temporal slots accumulate, each node interacts in with only a few nodes in a given temporal slot, the DCN can be represented by an \textit{High-Dimensional Sparse} (HDS) tensor. In order to extract rich behavioral patterns from an HDS tensor in DCN, this paper proposes an \textit{Adaptive Temporal-dependent Tensor low-rank representation} (ATT) model. It adopts a three-fold approach: a) designing a temporal-dependent method to reconstruct temporal feature matrix, thereby precisely represent the data by capturing the temporal patterns; b) achieving hyper-parameters adaptation of the model via the \textit{Differential Evolutionary Algorithms} (DEA) to avoid tedious hyper-parameters tuning; c) employing nonnegative learning schemes for the model parameters to effectively handle an the nonnegativity inherent in HDS data. The experimental results on four real-world DCNs demonstrate that the proposed ATT model significantly outperforms several state-of-the-art models in both prediction errors and convergence rounds.
\end{abstract}


\section{Introduction}

With the rapid advancement of the internet and communication technology~\cite{onnela2007analysis,deng2023bctc, zhou2023cryptocurrency}, the information transmission among a great number of communication devices such as computers, mobile devices, etc. generates intricate dynamic interactions, which can be effectively depicted by an \textit{Dynamic Communication Network}~\cite{yuan2022kalman} (DCN). Specifically, each node in a communication network denotes a communication device, links between nodes denote interaction behaviors, and weights on the links quantify interactions' strength. Further, the DCN is obtained by stacking the communication networks in different temporal slots chronologically, which contains rich behavioral patterns of the involved nodes~\cite{luo2021novel}. However, as the number of nodes increases dramatically and temporal slots accumulate heavily~\cite{li2022novel}, it is impossible for all nodes to build interaction behaviors in all temporal slots, and only partial node interactions are observed in few temporal slots, which results in DCNs that are \textit{High-Dimensional Sparse}~\cite{wu2021pid, luo2024pseudo, bi2023fast, luo2020position, wu2020data, luo2021fast, wu2023dynamic, liu2022symmetry} (HDS). Therefore, how to precisely extract the required knowledge from an HDS tensor becomes a tricky task.

Researchers have proposed a variety of different analysis approaches to a DCN for exploring behavioral and temporal patterns~\cite{lee2019dynamic, chen2022mnl, luo2019temporal, wu2022double, wu2022prediction, hu2020algorithm, hu2021effective, wei2022robust, xie2021acceleration, wang2021large, li2022using, qi2021robust}, such as \textit{Matrix Factorization} models, Ma \textit{et al.}~\cite{7831395} divide the tensor into different matrix slices and use respectively nonnegative matrix factorization to analyze the dynamic data; \textit{Graph Neural Network} (GNN) models~\cite{chen2024sdgnn}, Zhang \textit{et al}.~\cite{zhang2020relational} propose a relational GNN to analyze dynamic graphs by using neighborhood information; \textit{Recurrent Neural Network} (RNN) models, Jiao \textit{et al.}~\cite{jiao2021temporal} propose an RNN embedding method based on a variational auto-encoder; and \textit{Generative Adversarial Network} (GAN) models, Ren \textit{et al.}~\cite{ren2019fully} propose a fully data-driven GAN for dynamic data imputation. Although the above models are proven to be effective for the DCN analysis, their expensive computational and storage overheads limit their generalization to large-scale real-world applications.

\begin{figure}[t]
	\centerline{\includegraphics[width=0.95\textwidth]{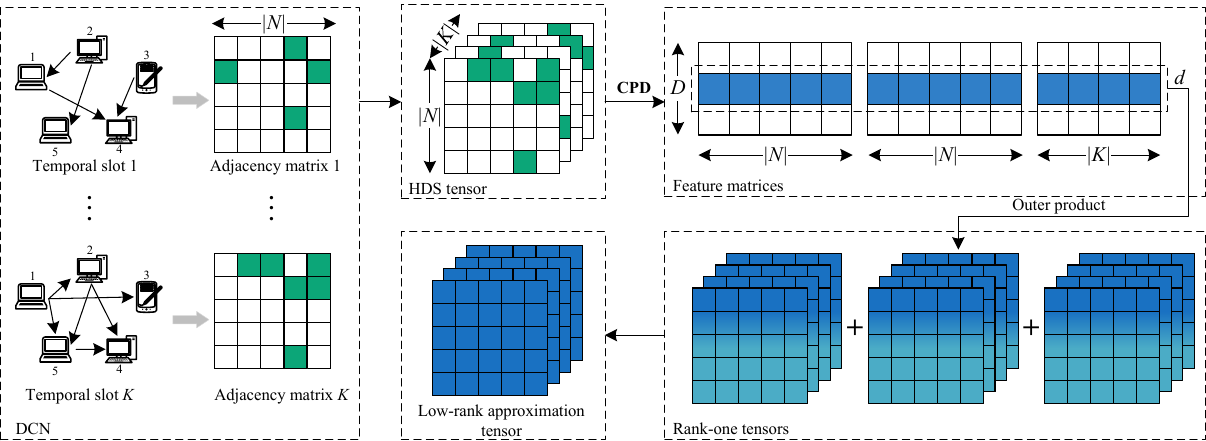}}
	\caption{A low-rank representation of HDS tensor constructed by a DCN.}
	\label{lft}
\end{figure}

Previous studies~\cite{luo2022neulft, qin2022low, luo2021adjusting, luo2021alternating, li2022momentum, bi2023two, yuan2024fuzzy, zhong2024alternating, jin2022neural, chen2022growing, xie2023biobjective} demonstrate that the \textit{Tensors Low-rank Representation} (TLR) models can precisely extract pattern information from HDS data by preserving the spatio-temporal structure of a DCN, thereby effectively performing behavioral and temporal pattern analysis. Therefore, this paper proposes an \textit{Adaptive Temporal-dependent Tensor low-rank representation} (ATT) model to perform DCNs analysis, which employs a \textit{Temporal-dependent Weight Matrix} (TWM) and a \textit{Differential Evolutionary Algorithm}~\cite{li2014differential, chen2022differential} (DEA) to provide excellent model performance. The main contributions of this paper are provided as follows:
\begin{itemize}
	\item A TWM-based approach. It considers the correlations between temporal slots to achieve lower prediction errors.
	\item A DEA-based hyper-parameters adaptation scheme. It automatically adjusts the best hyper-parameters during parameter learning to achieve fast model convergence.
	\item Extensive experimental evaluation on four real-world DCNs datasets. It indicates that the ATT model has better model performance in terms of prediction errors and  convergence rounds.
\end{itemize}

For the remained sections, Section~\ref{bg} presents the Background and Section~\ref{model} introduces the proposed ATT model in detail. In addition, Section\ref{es} shows the comparison experiments between ATT and competing models and Section~\ref{con} provides the conclusions and future works.

\section{Background}\label{bg}
\subsection{Symbol Definitions}
\begin{table}[t]
	\caption{The description for symbol definitions}
	\centering
	\begin{tabular}{@{}cl@{}}
		\toprule
		Symbol & Definition \\ \midrule
		$ N, K $ & The node and temporal slot sets. \\
		$\textbf{X} \in \mathbb{R}_{+}^{N \times N \times K}$ & The HDS tensor constructed via a DCN. \\
		$\tilde{\textbf{X}} \in \mathbb{R}_{+}^{N \times N \times K}$ & The low-rank approximation tensor of \textbf{X}.  \\ 
		$\textbf{R} \in \mathbb{R}_{+}^{N \times N \times K}$ & The rank-one tensor for $\tilde{\textbf{X}}$. \\
		$ x_{ijk}, \tilde{x}_{ijk}, r_{ijk} $  & The single element of $ \textbf{X}, \tilde{\textbf{X}} $, and $ \textbf{R} $. \\
		$ D $ & The number of rank-one tensors. \\
		$ \{\mathrm{S}, \mathrm{U}\} \in \mathbb{R}_{+}^{N \times D}$ & The node feature matrices \\
		$ \mathrm{Z} \in \mathbb{R}_{+}^{K \times D}$ & The temporal feature matrix \\
		$ \textbf{\textit{s}}_{d}, \textbf{\textit{u}}_{d}, \textbf{\textit{z}}_{d} $ & The $ d $-th column vectors of S, U, and Z. \\
		$ s_{id}, u_{jd}, z_{kd} $ & The single element of S, U, and Z. \\
		$ \mathrm{W} \in \mathbb{R}_{+}^{K \times K}$ & The TWM. \\
		
		$ \{\textbf{\textit{a}}, \textbf{\textit{c}}\} \in \mathbb{R}_{+}^{N}, \textbf{\textit{e}}\in \mathbb{R}_{+}^{K} $ & The node and temporal bias vectors \\
		$ a_{i}, c_{j}, e_{k} $ & The single element of $ \textbf{\textit{a}} $, $ \textbf{\textit{c}} $, and $ \textbf{\textit{e}} $. \\ \bottomrule
	\end{tabular}
	\label{symbol}
\end{table}

The symbol definitions of this paper are listed in Table~\ref{symbol}.

\subsection{Problem Formulation}

As illustrated in Fig.~\ref{lft}, the communication network for each temporal slot is represented by the corresponding adjacency matrix and a third-order tensor can be constructed by stacking the adjacency matrices along the temporal dimension.

\textbf{Definition 1:} Let $ \Omega $ and $ \Upsilon $ denote the observed and unobserved element sets for a tensor $ \textbf{X} $, respectively. If $ |\Upsilon|  \gg   |\Omega| $, $ \textbf{X} $ is an HDS tensor~\cite{wu2020advancing}, where each element $ x_{ijk} \in \textbf{X} $ denotes the interaction weight between the communication nodes $ i \in N $ and $ j\in N $ at the temporal slot $ k \in K $.

Therefore, following the principle of \textit{Canonical Polyadic Decomposition}~\cite{zhang2017improved} (CPD), the HDS tensor $ \textbf{X} $ is decomposed into three feature matrices $ \mathrm{S}, \mathrm{U} $, and $ \mathrm{Z} $ by the low-rank representation approach. With it, the low-rank approximation $ \tilde{\textbf{X}} $ of $\textbf{X}$ contains the $ D $ rank-one tensors as:
\begin{equation}
	\tilde{\textbf{X}} = \sum\limits_{d = 1}^D {{\textbf{R}_d}} ,
\end{equation}
where a rank-one tensor $ \textbf{R}_{d} $ is formed by the $ d $-th row vector of the three feature matrices $ \mathbf{R}_d= \textbf{\textit{s}}_d\circ \textbf{\textit{u}}_d\circ  \textbf{\textit{z}}_d $. Further, we can obtain a fine-grained single element $ \tilde{x}_{ijk} $ as following:
\begin{equation}
	{\tilde{x}_{ijk}} = \sum\limits_{d = 1}^D {r_{ijk}^{(d)}}  = \sum\limits_{d = 1}^D {{s_{id}}{u_{jd}}{z_{kd}}} .
\end{equation}

By previous studies~\cite{luo2019temporal, wang2024distributed, qin2023asynchronous, wu2023robust, wu2023mmlf, yuan2024adaptive, li2023generalized, liu2023symmetry, jin2021distributed, wei2022noise}, incorporating the corresponding bias $ \textbf{\textit{a}}, \textbf{\textit{c}} $, and $ \textbf{\textit{e}} $ can effectively suppress the fluctuation of HDS data over time. Hence, $ \tilde{x}_{ijk} $ is reformulated as below:
\begin{equation}
	{\tilde{x}_{ijk}} = \sum\limits_{d = 1}^D {{s_{id}}{u_{jd}}{z_{kd}}} + a_i + c_j + e_k.
	\label{3}
\end{equation}

In order to obtain the desired feature matrices and bias vectors, it is common to employ the Euclidean distance~\cite{dokmanic2015euclidean, qin2023adaptively, li2023saliency, luo2021fast, luo2023predicting, hu2023fcan, wu2023graph, qin2023parallel, wang2024dynamically, xiao2022novel} to model the gap between the original tensor $ \textbf{X} $ and the lower-order approximation tensor $ \tilde{\textbf{X}} $. Note that the original tensor is HDS, it is efficient to measure the gap on observed data $ \Omega $ in $ \textbf{X} $. Meanwhile, we adopt $ L_2 $ norm-based regularization~\cite{shah2016inverse, yang2023highly, li2022diversified, chen2023tensor,liu2023high, li2022second, xu2023hrst, yuan2023adaptive} to avoid model overfitting. Therefore, the objective function $ \varepsilon $ is constructed as follows:
\begin{equation}
	\begin{array}{l}
		\varepsilon  = \frac{1}{2}\sum\limits_\Omega  {{{\left( {{x_{ijk}} - \sum\limits_{d = 1}^D {{s_{id}}{u_{jd}}{z_{kd}}}  - {a_i} - {c_j} - {e_k}} \right)}^2}} \\
		+ \frac{1}{2}\sum\limits_\Omega  {{\left( {\lambda \sum\limits_{d = 1}^D {\left( {s_{id}^2 + u_{jd}^2 + z_{kd}^2} \right)}  + {\lambda _b}\left( {a_i^2 + c_j^2 + e_k^2} \right)} \right)}},
	\end{array}
\end{equation}
where $ \lambda $ and $ \lambda_b $ denote the regularization constant of controlling feature matrices and bias vectors, respectively. $ \Omega $ denote the observed data on HDS tensor $ \textbf{X} $. Moreover, since the HDS data in the DCN is nonnegative, it is necessary to incorporate constraints for feature matrices and bias vectors to accurately characterize the data's nonnegativity as:
\begin{equation}
	\begin{array}{l}
		\varepsilon  = \frac{1}{2}\sum\limits_\Omega  {{{\left( {{x_{ijk}} - \sum\limits_{d = 1}^D {{s_{id}}{u_{jd}}{z_{kd}}}  - {a_i} - {c_j} - {e_k}} \right)}^2}} \\
		+ \frac{1}{2}\sum\limits_\Omega  {{\left( {\lambda \sum\limits_{d = 1}^D {\left( {s_{id}^2 + u_{jd}^2 + z_{kd}^2} \right)}  + {\lambda _b}\left( {a_i^2 + c_j^2 + e_j^2} \right)} \right)}}, \\
		s.t.\forall i \in N,\forall j \in N,\forall k \in K,\forall d \in \{ 1 \sim D\} :\\
		{s_{id}} \ge 0,{u_{jd}} \ge 0,{z_{kd}} \ge 0,{a_i} \ge 0,{c_j} \ge 0,{e_k} \ge 0.
	\end{array}
\end{equation}

\begin{figure}[t]
	\centerline{\includegraphics[width=0.8\textwidth]{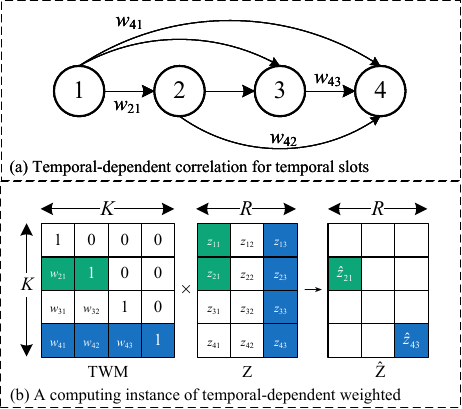}}
	\caption{An illustration of the temporal-dependent.}
	\label{td}
\end{figure}

\section{ATT model}\label{model}
\subsection{Objective Function}\label{of}

As mentioned previously, the DCN is time-varying in nature, and there is a specific influence between the temporal slots $ k $ and $ k+1 $, i.e., subsequent communication interactions are influenced by previous interaction situations, as shown in Fig.~\ref{td}(a). Therefore, we design a TWM $ \mathrm{W} $ to learn the correlation between temporal slots, which reacts to the learning process of the temporal feature matrix and bias vector. Fig.~\ref{td}(b) presents the computing instance between the temporal feature matrix and TWM, with the same computing for the temporal bias vector, they are given as:
\begin{equation}
	{\hat z_{kd}} = \sum\limits_{l = 1}^k {{w_{kl}}{z_{ld}}} ,\quad {\hat e_k} = \sum\limits_{l = 1}^k {{w_{kl}}{e_l}} .
\end{equation}
Note that for the learnable TWM, $ w_{kl} = 0, l > k $ since subsequent temporal information does not influence the previous interactions. Also, the diagonal weights are set to $ w_{kk}=1 $ since the temporal slot $ k $ has a constant correlation with itself. Therefore, the temporal-dependent objective function is reformulated as follows:

\begin{equation}
	\begin{array}{*{20}{l}}
		{\varepsilon  = \frac{1}{2}\sum\limits_\Omega  {{{\left( {{x_{ijk}} - \sum\limits_{d = 1}^D {{s_{id}}{u_{jd}}{{\hat z}_{kd}}}  - {a_i} - {c_j} - {{\hat e}_k}} \right)}^2}} }\\
		{ + \frac{1}{2}\sum\limits_\Omega  {\left( {\lambda \sum\limits_{d = 1}^D {\left( {s_{id}^2 + u_{jd}^2 + \hat z_{kd}^2} \right)}  + {\lambda _b}\left( {a_i^2 + c_j^2 + \hat e_k^2} \right)} \right)} ,}\\
		{s.t.\forall i \in N,\forall j \in N,\forall k \in K,\forall d \in \{ 1 \sim D\} :}\\
		{{s_{id}} \ge 0,{u_{jd}} \ge 0,{z_{ld}} \ge 0,{a_i} \ge 0,{c_j} \ge 0,{e_l} \ge 0,{w_{kl}} \ge 0.}
	\end{array}
\end{equation}

\subsection{Parameters Learning}\label{pl}

In order to learn the desired feature matrices, bias vectors, and learnable weight matrix, we firstly adopt the \textit{Additive Gradient Descent} (AGD) approach~\cite{liu2020convergence, li2023nonlinear, chen2024generalized, luo2021generalized} to implement their learning scheme\footnote{For brevity, we next present the inferences for $ s_{id}, z_{ld}, a_i, e_l $, and $ w_{kl} $, where the inferences of $ u_{jd} $ and $ c_j $ are similar to that of $ s_{id} $ and $ a_i $.} as:
\begin{equation}
	\begin{array}{l}
		\left( {\mathrm{S},\mathrm{U},\mathrm{Z},\textbf{\textit{a}},\textbf{\textit{c}},\textbf{\textit{e}},\mathrm{W}} \right) = \mathop {\arg \min }\limits_{\mathrm{S},\mathrm{U},\mathrm{Z},\textbf{\textit{a}},\textbf{\textit{c}},\textbf{\textit{e}},\mathrm{W}} \varepsilon \mathop  \Rightarrow \limits^{\mathrm{AGD}}   \\
		\left\{ \begin{array}{l}
			{s_{id}} \leftarrow {s_{id}} - {\eta _{id}}\frac{{\partial \varepsilon }}{{\partial {s_{id}}}} = {s_{id}}\\
			\quad \; - {\eta _{id}}\sum\limits_{{\Omega _{\left( i \right)}}} {\left( {\left( {{x_{ijk}} - {{\tilde x}_{ijk}}} \right)\left( { - {u_{jd}}{{\hat z}_{kd}}} \right) + \lambda {s_{id}}} \right)}; \\
			{z_{ld}} \leftarrow {z_{ld}} - {\eta _{ld}}\frac{{\partial \varepsilon }}{{\partial {{\hat z}_{kd}}}}\frac{{\partial {{\hat z}_{kd}}}}{{\partial {z_{ld}}}} = {z_{ld}}\\
			\quad \; - {\eta _{ld}}\sum\limits_{{\Omega _{\left( k \right)}}} {\left( {\left( {\left( {{x_{ijk}} - {{\tilde x}_{ijk}}} \right)\left( { - {s_{id}}{u_{jd}}} \right) + \lambda {{\hat z}_{kd}}} \right){w_{kl}}} \right)}; \\
			{a_i} \leftarrow {a_i} - {\eta _i}\sum\limits_{{\Omega _{\left( i \right)}}} {\left( { - \left( {{x_{ijk}} - {{\tilde x}_{ijk}}} \right) + {\lambda _b}{a_i}} \right)}; \\
			{e_l} \leftarrow {e_l} - {\eta _l}\sum\limits_{{\Omega _{\left( k \right)}}} {\left( {\left( { - \left( {{x_{ijk}} - {{\tilde x}_{ijk}}} \right) + {\lambda _b}{{\hat e}_k}} \right){w_{kl}}} \right)}; \\
			{w_{kl}} \leftarrow {w_{kl}} - {\eta _{kl}}\left( {\frac{{\partial \varepsilon }}{{\partial {{\hat z}_{kd}}}}\frac{{\partial {{\hat z}_{kd}}}}{{\partial {w_{kl}}}} + \frac{{\partial \varepsilon }}{{\partial {{\hat e}_k}}}\frac{{\partial {{\hat e}_k}}}{{\partial {w_{kl}}}}} \right) = {w_{kl}}\\
			\quad \; - {\eta _{kl}}\sum\limits_{{\Omega _{\left( k \right)}}} {\left( {\left( {\left( {{x_{ijk}} - {{\tilde x}_{ijk}}} \right)\left( { - {s_{id}}{u_{jd}}} \right) + \lambda {{\hat z}_{kd}}} \right){z_{ld}}} \right)} \\
			\quad \; - {\eta _{kl}}\sum\limits_{{\Omega _{\left( k \right)}}} {\left( {\left( { - \left( {{x_{ijk}} - {{\tilde x}_{ijk}}} \right) + {\lambda _b}{{\hat e}_k}} \right){e_l}} \right)},
		\end{array} \right.
	\end{array}
	\label{ls}
\end{equation}
where $ \eta_{id}$, $\eta_{ld}$, $\eta_i$, $\eta_{l} $, and $ \eta_{kl} $ are the learning rates of $ s_{id}, z_{ld}, a_i, e_{l} $, and $ w_{kl} $, respectively.  $ \Omega_{\left( i \right)} $ and $\Omega_{\left( k \right)}$ denote the subsets of $ \Omega $ linked with $ i \in N $ and $ k \in K $, respectively. Note that applying AGD approach directly doesn't satisfy the nonnegativity condition in the objective function. Therefore, we follow the principle of \textit{Nonnegative Multiplication Update}~\cite{song2022nonnegative,wu2019posterior, luo2021symmetric} (NMU) to adjust the learning rate as follows:
\begin{equation}
	\left\{ \begin{array}{l}
		{\eta _{id}} = {{{s_{id}}} \mathord{\left/
				{\vphantom {{{s_{id}}} {\sum\limits_{{\Omega _{\left( i \right)}}} {\left( {{{\tilde x}_{ijk}}{u_{jd}}{{\hat z}_{kd}} + \lambda {s_{ir}}} \right)} }}} \right.
				\kern-\nulldelimiterspace} {\sum\limits_{{\Omega _{\left( i \right)}}} {\left( {{{\tilde x}_{ijk}}{u_{jd}}{{\hat z}_{kd}} + \lambda {s_{ir}}} \right)} }};\\
		{\eta _{ld}} = {{{z_{ld}}} \mathord{\left/
				{\vphantom {{{z_{ld}}} {\sum\limits_{{\Omega _{\left( k \right)}}} {\left( {{{\tilde x}_{ijk}}{s_{id}}{u_{jd}} + \lambda {{\hat z}_{kd}}} \right){w_{kl}}} }}} \right.
				\kern-\nulldelimiterspace} {\sum\limits_{{\Omega _{\left( k \right)}}} {\left( {{{\tilde x}_{ijk}}{s_{id}}{u_{jd}} + \lambda {{\hat z}_{kd}}} \right){w_{kl}}} }};\\
		{\eta _i} = {{{a_i}} \mathord{\left/
				{\vphantom {{{a_i}} {\sum\limits_{{\Omega _{\left( i \right)}}} {\left( {{{\tilde x}_{ijk}} + {\lambda _b}{a_i}} \right)} }}} \right.
				\kern-\nulldelimiterspace} {\sum\limits_{{\Omega _{\left( i \right)}}} {\left( {{{\tilde x}_{ijk}} + {\lambda _b}{a_i}} \right)} }};\\
		{\eta _l} = {{{e_l}} \mathord{\left/
				{\vphantom {{{e_l}} {\sum\limits_{{\Omega _{\left( k \right)}}} {\left( {{{\tilde x}_{ijk}} + {\lambda _b}{{\hat e}_k}} \right){w_{kl}}} }}} \right.
				\kern-\nulldelimiterspace} {\sum\limits_{{\Omega _{\left( k \right)}}} {\left( {{{\tilde x}_{ijk}} + {\lambda _b}{{\hat e}_k}} \right){w_{kl}}} }};\\
		{\eta _{kl}} = {{{w_{kl}}} \mathord{\left/
				{\vphantom {{{w_{kl}}} {\sum\limits_{{\Omega _{\left( k \right)}}} {\left( {{{\tilde x}_{ijk}}{s_{id}}{u_{jd}} + \lambda {{\hat z}_{kd}}} \right){z_{ld}} + \left( {{{\tilde y}_{ijk}} + {\lambda _b}{{\hat e}_k}} \right){e_l}} }}} \right.
				\kern-\nulldelimiterspace} {\sum\limits_{{\Omega _{\left( k \right)}}} {\left( {{{\tilde x}_{ijk}}{s_{id}}{u_{jd}} + \lambda {{\hat z}_{kd}}} \right){z_{ld}} + \left( {{{\tilde y}_{ijk}} + {\lambda _b}{{\hat e}_k}} \right){e_l}} }}.
	\end{array} \right.
	\label{lr}
\end{equation}
By submitting the learning rates in \eqref{lr} to the learning scheme in \eqref{ls}, we can obtain the following nonnegative update rules:
\begin{equation}
	\left\{ \begin{array}{l}
		{s_{id}} \leftarrow {{{s_{id}}\sum\limits_{{\Omega _{\left( i \right)}}} {{x_{ijk}}{u_{jd}}{{\hat z}_{kd}}} } \mathord{\left/
				{\vphantom {{{s_{id}}\sum\limits_{{\Omega _{\left( i \right)}}} {{x_{ijk}}{u_{jd}}{{\hat z}_{kd}}} } {\sum\limits_{{\Omega _{\left( i \right)}}} {{{\tilde x}_{ijk}}{u_{jd}}{{\hat z}_{kd}} + \lambda {s_{id}}} }}} \right.
				\kern-\nulldelimiterspace} {\sum\limits_{{\Omega _{\left( i \right)}}} {{{\tilde x}_{ijk}}{u_{jd}}{{\hat z}_{kd}} + \lambda {s_{id}}} }};\\
		{z_{ld}} \leftarrow {{{z_{ld}}\sum\limits_{{\Omega _{\left( k \right)}}} {\left( {{x_{ijk}}{s_{id}}{u_{jd}}} \right)} } \mathord{\left/
				{\vphantom {{{z_{ld}}\sum\limits_{{\Omega _{\left( k \right)}}} {\left( {{x_{ijk}}{s_{id}}{u_{jd}}} \right)} } {\sum\limits_{{\Omega _{\left( k \right)}}} {\left( {{{\tilde x}_{ijk}}{s_{id}}{u_{jd}} + \lambda {{\hat z}_{kd}}} \right)} }}} \right.
				\kern-\nulldelimiterspace} {\sum\limits_{{\Omega _{\left( k \right)}}} {\left( {{{\tilde x}_{ijk}}{s_{id}}{u_{jd}} + \lambda {{\hat z}_{kd}}} \right)} }};\\
		{a_i} \leftarrow {{{a_i}\sum\limits_{{\Omega _{\left( i \right)}}} {{x_{ijk}}} } \mathord{\left/
				{\vphantom {{{a_i}\sum\limits_{{\Omega _{\left( i \right)}}} {{x_{ijk}}} } {\sum\limits_{{\Omega _{\left( i \right)}}} {{{\tilde x}_{ijk}} + {\lambda _b}{a_i}} }}} \right.
				\kern-\nulldelimiterspace} {\sum\limits_{{\Omega _{\left( i \right)}}} {{{\tilde x}_{ijk}} + {\lambda _b}{a_i}} }};\\
		{e_l} \leftarrow {{{e_l}\sum\limits_{{\Omega _{\left( k \right)}}} {{x_{ijk}}} } \mathord{\left/
				{\vphantom {{{e_l}\sum\limits_{{\Omega _{\left( k \right)}}} {{x_{ijk}}} } {\sum\limits_{{\Omega _{\left( k \right)}}} {\left( {{{\tilde x}_{ijk}} + {\lambda _b}{{\hat e}_k}} \right)} }}} \right.
				\kern-\nulldelimiterspace} {\sum\limits_{{\Omega _{\left( k \right)}}} {\left( {{{\tilde x}_{ijk}} + {\lambda _b}{{\hat e}_k}} \right)} }};\\
		{w_{kl}} \leftarrow {{{w_{kl}}\sum\limits_{{\Omega _{\left( k \right)}}} {\left( {{x_{ijk}}{s_{id}}{u_{jd}}} \right){z_{ld}} + {x_{ijk}}{e_l}} } \mathord{\left/
				{\vphantom {{{w_{kl}}\sum\limits_{{\Omega _{\left( k \right)}}} {\left( {{x_{ijk}}{s_{id}}{u_{jd}}} \right){z_{ld}} + {x_{ijk}}{e_l}} } {}}} \right.
				\kern-\nulldelimiterspace} {}}\\
		\quad \quad \;\;\;\sum\limits_{{\Omega _{\left( k \right)}}} {\left( {{{\tilde x}_{ijk}}{s_{id}}{u_{jd}} + \lambda {{\hat z}_{kd}}} \right){z_{ld}} + \left( {{{\tilde y}_{ijk}} + {\lambda _b}{{\hat e}_k}} \right){e_l}} .
	\end{array} \right.
	\label{update}
\end{equation}
With \eqref{update}, it is demonstrated that if the parameters of ATT are initialized to be nonnegative, then they remain nonnegative during the learning process, thereby satisfying the constraints in the objective function.

\subsection{The DEA-based Hyper-parameters Adaptation}

As presented in the previous section, the learning rule for the model parameters relies on two hyper-parameters $ \lambda $ and $ \lambda_b $, which significantly affect the effectiveness of parameters learning. How to choose the appropriate hyper-parameters is a crucial step. To avoid the time and computational overhead of manually adjusting the hyperparameters, we adopt the DEA to achieve hyper-parameters adaptation due to the simple structure and excellent global convergence characteristics of DEA. Specifically, we initialize $ P $ individuals as a swarm, where the $ p $-th individual is defined as a vector $ \textbf{\textit{v}}_{(p)}=(\lambda_{(p)}, \lambda_{b(p)}) $, $ p=1,...,P$, it's given as:
\begin{equation}
	{\textbf{\textit{v}}_{\left( p \right)}} = \left\{ \begin{array}{l}
		{\lambda _{\left( p \right)}} = {\lambda ^{\min }} + \theta \left( {{\lambda ^{\max }} - {\lambda ^{\min }}} \right);\\
		{\lambda _{b\left( p \right)}} = \lambda _b^{\min } + \theta \left( {\lambda _b^{\max } - \lambda _b^{\min }} \right),
	\end{array} \right.
	\label{init}
\end{equation}
where $ (\lambda ^{\min },\lambda ^{\max }) $ and $ (\lambda_b ^{\min },\lambda_b ^{\max }) $ are the lower and upper bounds of $ \lambda $ and $ \lambda_b $, respectively. $ \theta $ is a random value in the range $ [0,1] $. Following the principle of DEA, each individual is performed the mutation operation as follows:
\begin{equation}
	\textbf{\textit{v}}_{\left( p \right)}^{t + 1} = \boldsymbol{\tau}  + \mu\left( {\textbf{\textit{v}}_{\left( {r1} \right)}^t - \textbf{\textit{v}}_{\left( {r2} \right)}^t} \right),
\end{equation}
where $\boldsymbol{\tau}$ is the global best individual, $ \mu  $ is the scaling factor, $ r_1 $ and $ r_2 $ are two randomly selected individuals. Note that each individual is bounded via \eqref{b} as:
\begin{equation}
	{\textbf{\textit{v}}^{t+1}_{\left( p \right)}} = \left\{ \begin{array}{l}
		{\lambda ^{\min }} \le {\lambda^{t+1} _{\left( p \right)}} \le {\lambda ^{\max }};\\
		\lambda _b^{\min } \le {\lambda^{t+1} _{b\left( p \right)}} \le \lambda _b^{\max }.
	\end{array} \right.
	\label{b}
\end{equation}

After that, each individual is performed the crossover operation as below:
\begin{equation}
	\textbf{\textit{v}}_{m\left( p \right)}^{t + 1} = \left\{ \begin{array}{l}
		\textbf{\textit{v}}_{m\left( p \right)}^{t + 1};\quad {\rm{if}}\;\theta  \le {C_p}\;{\rm{or}}\;m = {m^*}\\
		\textbf{\textit{v}}_{m\left( p \right)}^t,\quad {\rm{else}}
	\end{array} \right.
\end{equation}
where $ m \in \mathbb{N} $ denotes one of the dimensions in vector $ \textbf{\textit{v}}_{(p)} $, $ C_p \in [0,1] $ denotes the crossover probability, and $ m^{*} $ denotes a randomly selected dimension. Next, each individual is evaluated for fitness by the learning process of the model parameters, and the fitness function $ F $ for the $ t $-th iteration is defined as:
\begin{equation}
	F_{(p)}^{t+1} = \frac{{H\left( {\textbf{\textit{v}}_{(p)}^{t+1}} \right) - H\left( {\textbf{\textit{v}}_{(p - 1)}^{t+1}} \right)}}{{H\left( {\textbf{\textit{v}}_{(P)}^{t+1}} \right) - H\left( {\textbf{\textit{v}}_{(P)}^{t}} \right)}},
	\label{16}
\end{equation}
where $ H(\textbf{\textit{v}}^{t+1}_{(0)}) = H(\textbf{\textit{v}}^{(t)}_{P}) $ and $ H $ is calculated by the current performance of the model parameters, it's given as:
\begin{equation}
	H\left( {{\textbf{\textit{v}}_{(p)}}} \right) = \sqrt {\frac{{\sum\limits_\Lambda  {{{\left( {{x_{ijk}} - {{\tilde x}_{ijk\left( p \right)}}} \right)}^2}} }}{{4\left| \Lambda  \right|}}}  + \;\frac{{\sum\limits_\Lambda  {{{\left| {{x_{ijk}} - {{\tilde x}_{ijk\left( p \right)}}} \right|}_a}} }}{{2\left| \Lambda  \right|}},
	\label{h}
\end{equation}
where $ \Lambda $ denotes the validation set form $ \Omega $ and $ |\cdot|_{a} $ denotes the absolute value. Note that the approximation $ \tilde{x}_{ijk} $ in \eqref{h} is calculated via the current model parameters.  takes $ (s_{id}, a_{i}) $ and as an example, their learning rules are as follows:
\begin{equation}
	\left\{ \begin{array}{l}
		{s_{id\left( p \right)}} \leftarrow \frac{{{s_{id\left( p \right)}}\sum\limits_{{\Omega _{\left( i \right)}}} {{x_{ijk}}{u_{jd}}{{\hat z}_{kd}}} }}{{\sum\limits_{{\Omega _{\left( i \right)}}} {{{\tilde x}_{ijk\left( p \right)}}{u_{jd\left( p \right)}}{{\hat z}_{kd\left( p \right)}} + {\lambda _{\left( p \right)}}{s_{id\left( p \right)}}} }};\\
		{a_{i\left( p \right)}} \leftarrow \frac{{{a_{i\left( p \right)}}\sum\limits_{{\Omega _{\left( i \right)}}} {{x_{ijk}}} }}{{\sum\limits_{{\Omega _{\left( i \right)}}} {{{\tilde x}_{ijk\left( p \right)}} + {\lambda _{b\left( p \right)}}{a_{i\left( p \right)}}} }}.
	\end{array} \right.
\end{equation}

By the evaluation of fitness functions, the global best individual is updated as below:
\begin{equation}
	{\boldsymbol{\tau} ^{t + 1}} = \left\{ {\begin{array}{*{20}{c}}
			{{\boldsymbol{\tau} ^t},F_{(p)}^{t + 1} \le F_{(p - 1)}^{t + 1},}\\
			{\textbf{\textit{v}}_{(p)}^{t + 1},F_{(p)}^{t + 1} > F_{(p - 1)}^{t + 1}.}
	\end{array}} \right.
	\label{19}
\end{equation}
With \eqref{init}-\eqref{19}, we employ DEA to achieve hyper-parameters adaption of the model. And the specific process of the model. The specific process of the model is presented in Algorithm~\ref{Alg1}.

\begin{algorithm}[t]
	\caption{ATT model}
	\label{Alg1}
	Initialize input: $\Omega$, $N$, $K$, $D$, $ \textbf{\textit{v}} $, $ P $, $ T $ \;	
	Initialize model parameters: $ \mathrm{S} $, $ \mathrm{U} $, $ \mathrm{Z} $, $ \textbf{\textit{a}} $, $ \textbf{\textit{c}} $, $ \textbf{\textit{e}} $, $ \mathrm{W} $ with positive values\;
	\For{each $ x_{ijk} \in \Omega $}
	{
		Compute $\hat{x}_{ijk}$ according to \eqref{3}\;
	}
	Compute $ H(\textbf{\textit{v}}^0_{(P)}) $ according to \eqref{h}\;
	\While{$ t \le T$ \rm{and not converge}}
	{
		\ForEach{$ p = 1 $ to $ P $}
		{
			Update $ s_{id(p)} $, $ u_{jd(p)} $, $ z_{ld(p)} $, $ a_{i(p)} $, $ c_{j(p)} $, $ e_{l(p)} $, $ w_{kl(p)} $ according to \eqref{update}\;
		}
		Compute $ H(\textbf{\textit{v}}^t_{(p)}) $ according to \eqref{h}\;
		\ForEach{$ p = 1 $ to $ P $}
		{
			Compute $ F(\textbf{\textit{v}}^t_{(p)}) $ according to \eqref{16}\;
			update $ \boldsymbol{\tau} $ according to \eqref{19}\;
		}
		$ t = t + 1 $\;
	}

\end{algorithm}

\section{Empirical Studies}\label{es}
\subsection{Datasets}
\begin{table}[h]
	\caption{Dataset Details\label{t4.1}}
	\centering
	\begin{tabular}{ccccc}
		\toprule
		\textbf{Datasets} & \textbf{Nodes} & \textbf{Temporal Slots} & \textbf{Density} & \textbf{Observed}\\
		\midrule
		\textbf{D1} & 40072 & 318 & 4.82 $ \times 10^{-8} $ & 24638  \\
		\textbf{D2} & 46468 & 412 & 1.69 $ \times 10^{-8} $  & 15114\\
		\textbf{D3} & 49918 & 516 & 1.45 $ \times 10^{-8} $  & 18760 \\
		\textbf{D4} & 60114 & 618 & 2.54 $ \times 10^{-8} $ & 56826 \\
		\bottomrule
	\end{tabular}
\end{table}

In this paper, we use four real-world DCNs as evaluation datasets. Taking D1 as an example, nodes denotes 40072 communication nodes and temporal slots denotes that the dataset contains the communications of 318 time points. In addition, the number of observed data is 24638, which is 4.28$ \times 10^{-8} $ of the dataset. We divide the dataset into training set $ (\Gamma) $, validation set $ (\Lambda) $, and testing set $ (\Phi) $ in the ratio of 7:1:2 respectively. During the experiment, we repeat the division process 20 times to obtain unbiased  average results.

\subsection{Evaluation Metrics}

Based on previous studies, we employ the commonly utilized \textit{Root Mean Squared Error} (RMSE) and \textit{Mean Absolute Error}~\cite{wu2023robust, shi2020large, yuan2020multilayered, shang2021alpha, cheng2021novel, wu2021latent, hu2021fast, wu20211} (MAE) as evaluation metrics. Note that smaller RMSE and MAE indicate lower prediction error i.e. better model performance. They are given as below:
\begin{equation}
	{\rm{RMSE}} = \sqrt {\frac{{\sum\limits_\Phi  {{{\left( {{x_{ijk}} - {{\tilde x}_{ijk}}} \right)}^2}} }}{{\left| \Phi  \right|}}} {\rm{,}}\quad {\rm{MAE}} = \frac{{\sum\limits_\Phi  {{{\left| {{x_{ijk}} - {{\tilde x}_{ijk}}} \right|}_a}} }}{{\left| \Phi  \right|}}.
\end{equation}

\subsection{Training settings}

In order to fairly compare the performance of tested models, we uniformly set the rank for all models to 20~\cite{chen2021hierarchical, wang2022multi, hu2021distributed}. Note that the model training process is terminated if the number of training rounds reaches $ 10^{3} $ or the error between two consecutive training rounds is less than $ 10^{-5} $.

\subsection{Comparison Results}

\begin{table}[t]
	\caption{The prediction error for five tested models}
	\begin{center}
		\begin{tabular}{@{}ccccccc@{}}
			\toprule
			\textbf{Dataset} &  & \textbf{M1} & \textbf{M2} & \textbf{M3} & \textbf{M4} & \textbf{M5} \\ \midrule
			\multirow{2}{*}{\textbf{D1}} & \textbf{RMSE} & \textbf{0.2770} & 0.2836 & 0.2844 & 0.3007 & 0.2851 \\
			& \textbf{MAE} & \textbf{0.1899} & 0.2032 & 0.1971 & 0.2073 & 0.2031 \\
			\multirow{2}{*}{\textbf{D2}} & \textbf{RMSE} & \textbf{0.2444} & 0.2594 & 0.2639 & 0.2857 & 0.2815 \\
			& \textbf{MAE} & \textbf{0.1686} & 0.1828 & 0.1694 & 0.1981 & 0.1959 \\
			\multirow{2}{*}{\textbf{D3}} & \textbf{RMSE} & \textbf{0.2504} & 0.2624 & 0.2740 & 0.2821 & 0.2729 \\
			& \textbf{MAE} & \textbf{0.1643} & 0.1783 & 0.1675 & 0.1931 & 0.1862 \\
			\multirow{2}{*}{\textbf{D4}} & \textbf{RMSE} & \textbf{0.2645} & 0.2807 & 0.2734 & 0.2993 & 0.2830 \\
			& \textbf{MAE} & \textbf{0.1769} & 0.1958 & 0.1842 & 0.1995 & 0.1954 \\ \bottomrule
		\end{tabular}
		\label{tp}
	\end{center}
\end{table}

\begin{table}[t]
	\caption{The convergence rounds for five tested models}
	\begin{center}
		\begin{threeparttable}
			\begin{tabular}{@{}ccccccc@{}}
				\toprule
				\textbf{Dataset} &  & \textbf{M1} & \textbf{M2} & \textbf{M3} & \textbf{M4} & \textbf{M5} \\ \midrule
				\multirow{2}{*}{\textbf{D1}} & \textbf{CR-R$ ^* $} & 9$\pm$1 & 21$\pm$1 & 40$\pm$3 & 37$\pm$4 & 47$\pm$4 \\
				& \textbf{CR-M} & 23$\pm$1 & 26$\pm$1 & 47$\pm$3 & 40$\pm$2 & 49$\pm$5 \\
				\multirow{2}{*}{\textbf{D2}} & \textbf{CR-R} & 16$\pm$1 & 18$\pm$2 & 47$\pm$2 & 49$\pm$1 & 59$\pm$3 \\
				& \textbf{CR-M} & 18$\pm$1 & 27$\pm$2 & 61$\pm$5 & 48$\pm$3 & 64$\pm$5 \\
				\multirow{2}{*}{\textbf{D3}} & \textbf{CR-R} & 22$\pm$2 & 23$\pm$2 & 45$\pm$3 & 52$\pm$3 & 47$\pm$4 \\
				& \textbf{CR-M} & 20$\pm$3 & 29$\pm$3 & 72$\pm$6 & 53$\pm$3 & 60$\pm$3 \\
				\multirow{2}{*}{\textbf{D4}} & \textbf{CR-R} & 15$\pm$1 & 22$\pm$2 & 29$\pm$2 & 40$\pm$3 & 67$\pm$3 \\
				& \textbf{CR-M} & 22$\pm$1 & 27$\pm$2 & 29$\pm$2 & 41$\pm$2 & 66$\pm$3 \\ \bottomrule
			\end{tabular}
			\begin{tablenotes}
				\item[$ * $] CR-R denotes the Convergence Rounds for RMSE and CR-M for MAE.
			\end{tablenotes}
		\end{threeparttable}
		\label{tc}
	\end{center}
\end{table}

In this section, we use the ATT model (M1) to compare with four state-of-the-art models on four datasets in terms of prediction errors and convergence rounds as:
\begin{itemize}
	\item M2~\cite{luo2019temporal}: A biased tensor factorization with the NMU algorithm.
	\item M3~\cite{ye2021outlier}: A robust tensor factorization model with the Cauchy Loss function.
	\item M4~\cite{wang2016multi}: A integrated multi-linear algebraic model with the reconstructive optimization algorithm.
	\item M5~\cite{su2021tensor}: A multi-dimensional tensor factorization model with the alternating least square algorithm.
\end{itemize}
Table~\ref{tp} and Table~\ref{tc} show the statistics of all the models respectively. With these results, it can be seen that:
\begin{itemize}
	\item \textit{The ATT model exhibits lower prediction errors compared with the competing models}. As shown in Table~\ref{tp}, the ATT model achieves RMSE and MAE values of 0.2770 and 0.1899 on D1, respectively. In contrast, the competing models M2, M3, M4, and M5 have RMSE values of 0.2836, 0.2844, 0.3007, and 0.2851, and MAE values of 0.2032, 0.1971, 0.2073, and 0.2031, respectively. The prediction errors of ATT compared with that of the competing models in terms of RMSE and MAE are reduced by 2.30\%, 2.60\%, 7.88\%, 2.84\%, and 6.54\%, 3.65\%, 8.39\%, 6.49\%, respectively. In D2, the prediction errors of the ATT model, compared with the competing models, are reduced by 5.78\%, 7.38\%, 14.45\%, 13.17\%, and 7.76\%, 0.47\%, 14.89\%, 13.93\% in terms of RMSE and MAE, respectively.The similar performance enhancements can be observed in D3 and D4.
	
	\item \textit{The ATT model presents fewer convergence rounds compared with the competing models}. For instance, according to Table~\ref{tc}, the ATT model requires only 22 training rounds for RMSE and 20 training rounds for MAE to converge on D3. It's significantly fewer than the convergence rounds for M2's 23, M3's 45, M4's 52 and M5's 47 for RMSE, and M2's 29, M3's 72, M4's 53, and M5's 60 for MAE. In comparison, the convergence rounds of the ATT model for RMSE are 95.65\% of M2, 48.88\% of M3, 41.50\% of M4, and 46.80\% of M5, while for MAE, they are 68.96\% of M2, 27.77\% of M3, 37.73\% of M4, and 33.33\% of M5. On D4, the ATT model's convergence rounds for RMSE is 15, which is 68.18\%, 51.72\%, 37.50\%, and 22.38\% of the convergence rounds for the competing models. For MAE, the ATT model's convergence rounds is 81.48\%, 75.86\%, 53.65\%, and 33.33\% of the competing models. Similar results are observed on other datasets.
\end{itemize}

\section{Conclusions and future works}\label{con}

To achieve the analysis for LDCNs, this paper proposes an ATT model with lower prediction errors and fewer convergence rounds. In the ATT model, we design a TWM to capture the correlation of communication interactions at each time point, thereby representing the LDCNs more accurately. In addition, we implement the hyper-parameters adaptation of the ATT model to reduce the costly time and computational overhead for manual hyper-parameters tuning. Experiments on four real LDCNs demonstrate that the proposed ATT model can better analyze HDS data since the ATT model is superior to the competing models in terms of prediction errors and convergence rounds. However, the following two issues need to be addressed in future work:
\begin{itemize}
	\item The temporal-dependent correlations in this paper are modeled and learned by default as linear relationships, can we model temporal-dependent by employing nonlinear relationships~
\cite{xie2020rnn, qi2021recurrent, yan2023modified}?
	\item Can we use other learning schemes such as \textit{Augmented Lagrangian Method}~\cite{bi2023proximal} (ALM) to learn the model parameters?
\end{itemize}
We plan to study the above issues later.

\bibliographystyle{IEEEtran}
\bibliography{ATTbib}

\vspace{12pt}

\end{document}